\pdfoutput=1
\documentclass{article}
\usepackage{ijcai19}

\usepackage{times}
\usepackage{soul}
\usepackage{url}
\usepackage[utf8]{inputenc}
\usepackage[font=small]{caption}
\usepackage{graphicx}
\usepackage{amsmath}
\DeclareMathOperator{\argmin}{\mathrm{argmin}}
\usepackage{booktabs}
\usepackage{algorithm}
\usepackage{algorithmic}
\usepackage{amssymb}
\usepackage{hyperref} 
\usepackage{balance}
\usepackage[cjk]{kotex}

\usepackage{microtype}
\usepackage{makecell}%To keep spacing of text in tables
\setcellgapes{4pt}%parameter for the spacing
\usepackage{array,multirow}

\title{PuVAE: A Variational Autoencoder to Purify Adversarial Examples}

\author{
Uiwon Hwang$^{1,}$\thanks{Equal contribution}
\and
Jaewoo Park$^{1,*}$\and
Hyemi Jang$^{1,*}$\and
Sungroh Yoon$^{1,}$\thanks{Corresponding authors}\And
Nam Ik Cho$^{1,\dagger}$
\affiliations
$^1$Department of Electrical and Computer Engineering, Seoul
National University, Seoul, Korea.
\emails
$^{\dagger}$\{sryoon, nicho\}@snu.ac.kr
}

\begin{document}

%\footnote[1]{Corresponding Authors}

\maketitle
\begin{abstract}
Deep neural networks are widely used and exhibit excellent performance in many areas. However, they are vulnerable to adversarial attacks that compromise the network at the inference time by applying elaborately designed perturbation to input data. Although several defense methods have been proposed to address specific attacks, other attack methods can circumvent these defense mechanisms. Therefore, we propose Purifying Variational Autoencoder (PuVAE), a method to purify adversarial examples. The proposed method eliminates an adversarial perturbation by projecting an adversarial example on the manifold of each class, and determines the closest projection as a purified sample. We experimentally illustrate the robustness of PuVAE against various attack methods without any prior knowledge. In our experiments, the proposed method exhibits performances competitive with state-of-the-art defense methods, and the inference time is approximately 130 times faster than that of Defense-GAN that is the state-of-the art purifier model.
%Although deep neural networks have been used with excellent performance in various areas, it is vulnerable to adversarial attacks that fool the model at the inference time by applying elaborately designed perturbation to input data. Several methods have been proposed to address specific attacks, but other attack methods can circumvent these defense mechanisms. On the other hand, some defense mechanisms aim to eliminate adversarial perturbation by projecting an adversarial example onto the range of true data distribution. Since the latter approach can provide generalized defense ability, we propose a generative model to \textit{purify} adversarial examples, Purifying Variational Autoencoder (PuVAE). We succeed to purify various attack methods, and effectively reduce the time for generating purified sample compared to Defense-GAN.
\end{abstract}

\section{Introduction} %의원
\begin{figure}[t]
\centering
\includegraphics[width=0.8\columnwidth]{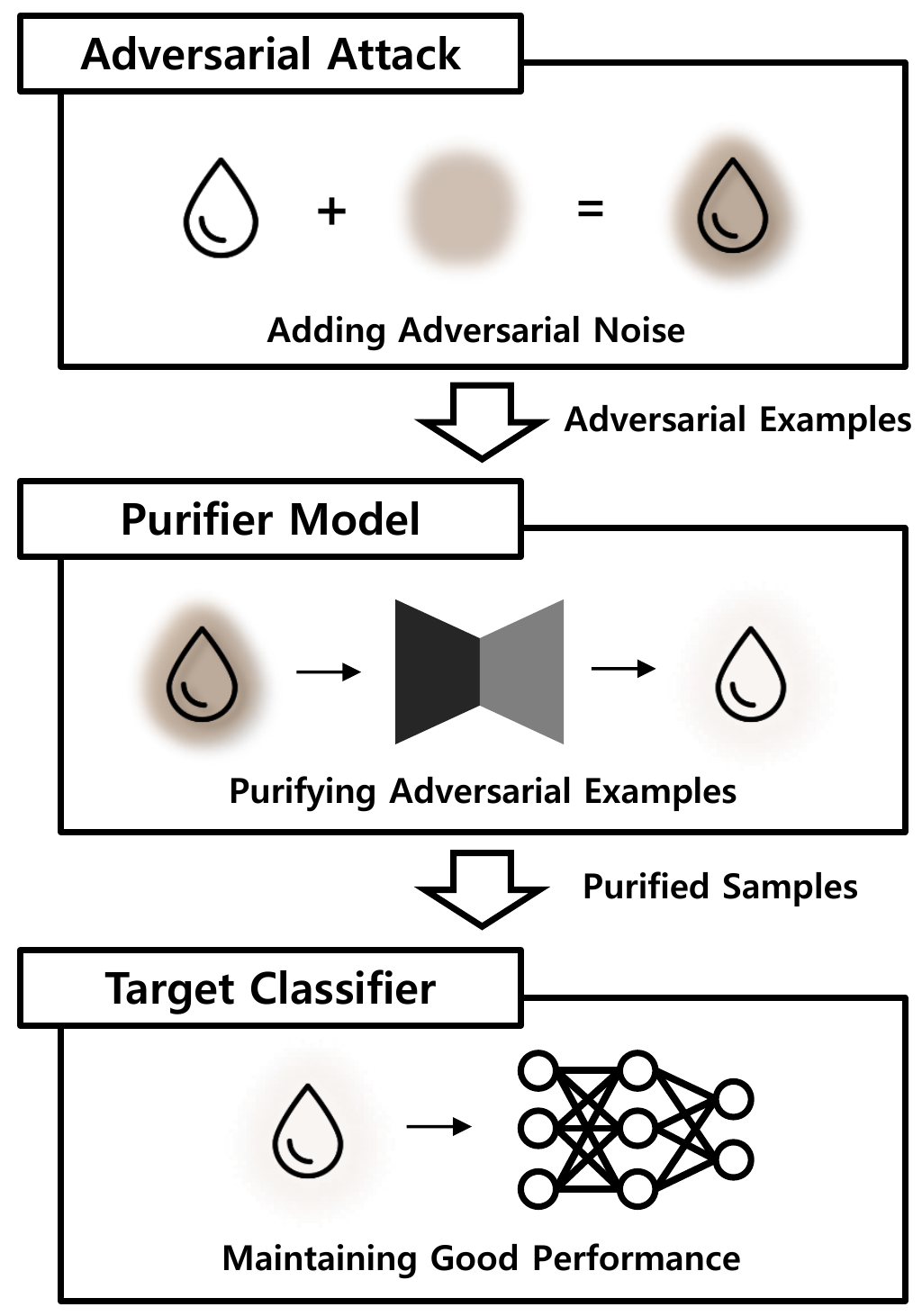} \caption{Overview of the defense mechanism using the purifier model.}
\label{fig:overview}
\end{figure}

% deep neural network가 많은 분야에서 우수한 성능으로 활용되어오고 있지만, perturbation을 통해 DNN의 학습을 완전히 망쳐놓는 adversarial example에 취약하다. 
Significant progress has characterized deep learning in several areas including image recognition \cite{resnet}, disease prediction \cite{hwang}, and autonomous driving \cite{wodbs}. However, security issues of deep neural networks, which are especially vulnerable to adversarial attacks, are emerging. The goal of adversarial attacks is to fool deep neural networks via applying elaborately designed perturbation to input data. Adversarial attacks make it hazardous to apply deep neural networks in real world applications. In the case of autonomous driving \cite{autonomous_adv}, attacks can cause an accident by making an object detector recognize pedestrians as roads.

\begin{figure}[t]
\includegraphics[width=1\columnwidth]{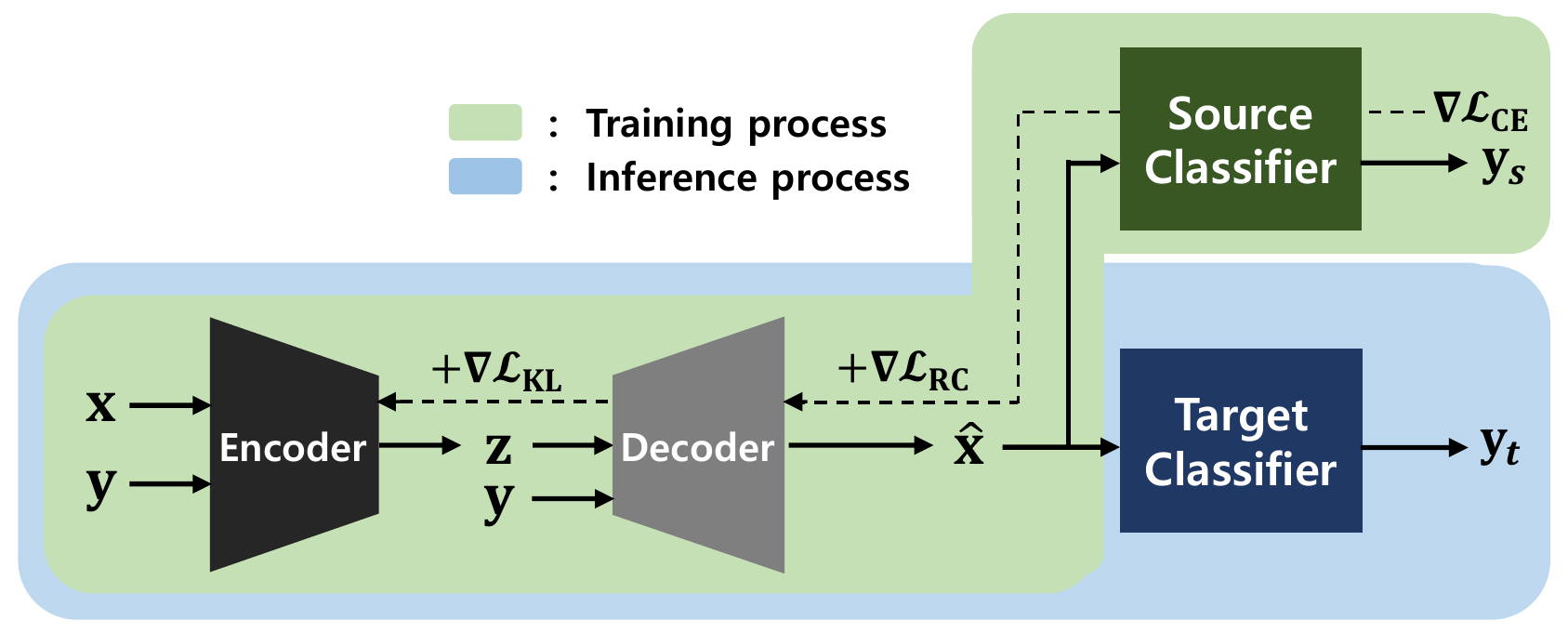} 
\caption{Overview of the PuVAE algorithm; The green region represents the training process, and the blue region denotes the inference process of PuVAE. The dotted line is the gradient flow in training process. The parameters of the source classifier are not updated.}
\label{fig:model}
\end{figure}

\begin{figure*}[t]
\centering
\includegraphics[width=0.9\textwidth]{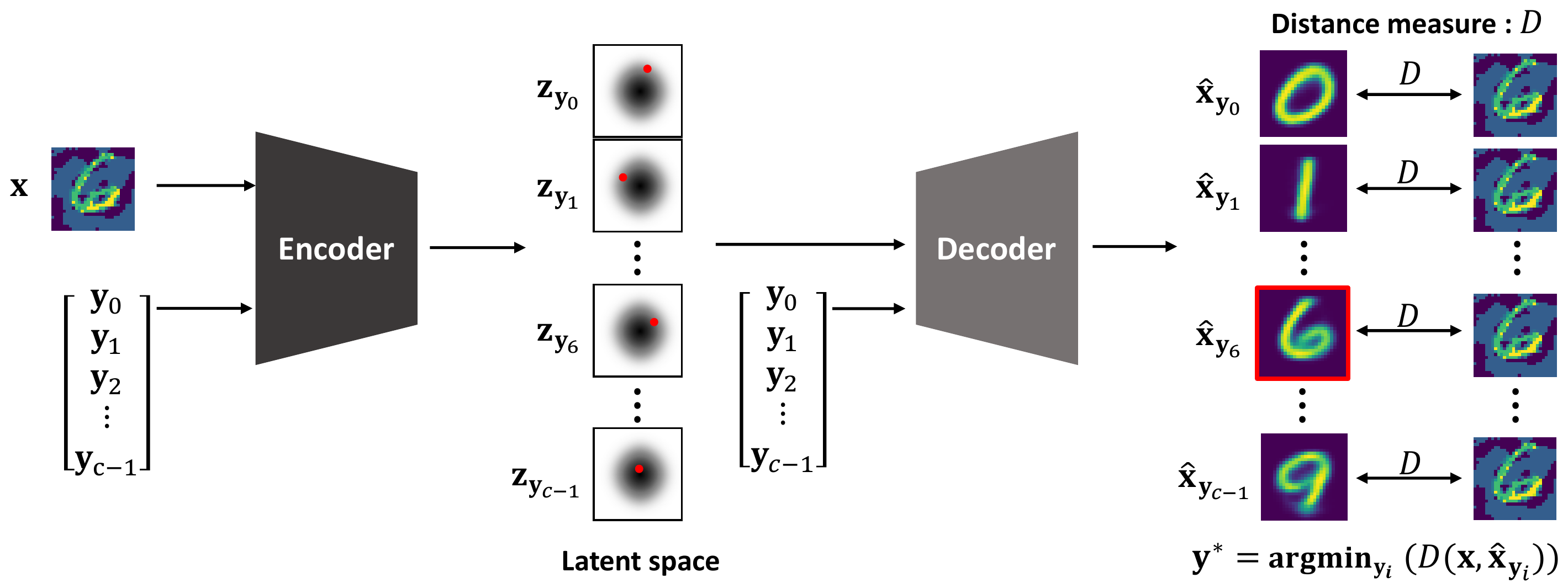} 
\caption{Inference of PuVAE with an MNIST image; A latent vector $\mathbf{z}_{\mathbf{y}_i}$ is sampled from an adversarial sample $\mathbf{x}$ and a class condition $\mathbf{y}_i$.}
\label{fig:PuVAE_model}
\end{figure*}

To address these attacks, several defense mechanisms have been proposed. There are three categories of defense mechanisms. The first mechanisms involve modifying the training dataset such that the classifier is robust against the adversarial attack \cite{szegedy2013intriguing}. Second mechanisms block gradient calculation via changing the training procedure \cite{buckman2018thermometer,guo2017countering}. However, they are only effective for the gradient based attack methods. The third mechanisms involve removing the adversarial noise from the sample fed into the classifier \cite{defensegan}.

% 우리는 이 중 adversarial perturbation이 더해졌을지도 모르는 input data를 정화하는 방법들에 흥미를 느꼈다, in that white-box와 black-box attack에 모두 효과적으로 대처할 수 있다는 점에서. 이 방법들은 generative model을 이용하여 prior distribution을 학습한 뒤 adversarial example이 들어오면 이를 학습된 p(x)로 project하는 방식으로 작동한다. 우리는 이러한 generative models를 purifiers라고 부르며, 여기에 대한 최근 work로 MagNet~\ref{}과 Defense-GAN~\ref{}이 있다. 

Our main focus is defense mechanisms to purify input data that may have added adversarial perturbation; this can allow the mechanisms to effectively address any attacks. These methods mostly work by using a generative model to learn the data distribution and project the adversarial example into the learned data distribution $p(x)$. We term the generative models as \textit{purifiers}, and MagNet~\cite{magnet} and Defense-GAN~\cite{defensegan} are recent work. In Figure~\ref{fig:overview}, we show an overview of the defense mechanism using the purifier model.

% MagNet은 reformer network라고 불리는 하나 또는 여러개의 autoencoder를 이용하여 original data를 학습한 뒤, input data를 autoencoder에 통과시켜 manifold에 가깝게 옮긴다. 그 결과 나온 output을 classifier에 공급한다. 하지만 이 방법은 defense-GAN에 비해 성능이 좋지 못하다는 단점이 있다.
Specifically, MagNet learns the original data using one or more autoencoders termed as the reformer networks and passes input data to the autoencoders that move input data closer to the data manifold. The purified data are supplied to the classifier. However, the method has a disadvantage wherein it exhibits poor performance when compared with Defense-GAN.

% Defense-GAN은 GAN의 특성을 활용하여 adversarial attack을 방어한다. GAN의 objective function을 optimize하는 것이 p_g와 p_data를 같게 만드는 것과 equivalent하다는 점을 이용하였다. original data를 이용하여 GAN을 학습한 뒤 L steps of GD로 x와 G(z)간의 reconstruction error를 줄이는 z를 gradient를 통해 iterative하게 찾아 G(z)를 classifier에게 공급한다. Defense-GAN은 현재까지 제안된 defense mechanism의 state-of-the art 이지만, 불안정한 GAN의 학습에 의존하고 adversarial noise까지 재현할 수 있다는 점에서 개선의 여지가 남아있다.
Defense-GAN uses the characteristics of generative adversarial networks (GANs) to defend a target model against adversarial attacks. It uses the fact that optimizing the objective function of the GAN is equivalent to making the generator distribution $p_{g}$ identical to the data distribution $p_{data}$. After training the GAN using the original data, Defense-GAN iteratively finds generator input $z$ through the gradients. This reduces the reconstruction error between the generated data $G(z)$ and the input data $x$ that may have added adversarial noise. Subsequently, data generated with optimal $z$ are supplied to the classifier as input. The model relies on the unstable performance of GAN, and this occationally reproduces adversarial noise by directly optimizing errors between the adversarial example $x+\delta$ and the generated sample $G(z)$. In addition, because of Defense-GAN's iterative nature, it takes a long time to yield the maximum defense performance. In particular, real-time applications such as object detection must operate in a short period of time, so a fast defense algorithm needs to be developed.

% 본 논문에서 우리는 adversarial examples로부터 well-classified samples를 생성하는 것을 목표한다. 이러한 '정화된' sample들은 target classifier로 공급되어 adversarial attack의 영향을 받지 않고 학습되도록 한다. MagNet과 Defense-GAN의 한계점을 극복하기 위해, 우리는 VAE를 이용한 adversarial examples를 정화하는 PuVAE와 saliency map을 이용해 purifier에 더 많은 정보를 제공하는 masked PuVAE를 제안한다. 제안된 모델은 amortized variational inference를 이용하여 MagNet보다 더 강인한 sample을 생성하고, adversarial example과의 직접적인 loss가 없어 adversarial noise를 재현하지 않도록 설계되었다. 이를 통해 우리는 제안된 모델이 기존 모델들보다 classifier를 위해 더 좋은 sample을 생성하는 purifier로서 동작하기를 기대한다.
In this paper, we aim to rapidly generate well-classified samples from adversarial examples. The \textit{purified} samples are fed into the target classifier so that they are classified without being affected by adversarial attacks. To solve the limitations of MagNet and Defense-GAN, we propose Purifying Variational AutoEncoder (PuVAE) that purifies adversarial examples using a Variational Autoencoder (VAE). The proposed model uses variational inference to generate samples that provide comparable or better defense performance than the state-of-the-art model. In contrast to Defense-GAN, PuVAE generates clean samples with one feed-forward step. Therefore, our method is robust against adversarial attacks within a reasonable time limit.

In summary, our contributions are as follows:
\begin{itemize}
    \item We propose a VAE-based defense method, PuVAE, to effectively purify adversarial attacks. The proposed method shows a remarkable performanc over other defense methods.
    \item The proposed method significantly reduces the time to generate purified samples. Within a reasonable time limit, PuVAE outperforms state-of-the-art defense methods. 
    \item Experimental results demonstrate that our method functions robustly against a variety of attack methods and datasets.
\end{itemize}
%we expect that the proposed models will act as a robust purifier to generate better samples for classifier than previous models. 

% In this paper, our goal is to make well-classified samples from adversarial examples, and convey it to the target model, a classifier. To make purifier, we propose two methods, Purifying-VAE (PuVAE) and Masked Purifying-VAE (Masked PuVAE). First, Purifying-VAE generates normal data from adversarial examples based on the normal data distribution learned during training time. Second, Distance-VAE use conditional VAE as a purifier. 

\section{Background}

\subsection{Variational Autoencoder (VAE)} % 재우
% 길면 네줄 지워도됨
A generative model is used to represent data distribution. Most data are too complex to directly find the relation in themselves, and thus relatively simple latent variables are typically used to represent data distribution.
Kingma~\shortcite{Kingma2013variational} introduced Variational AutoEncoder (VAE), which is a method that uses a combination of neural networks and variational inference to learn a decoder that generates data from normal distribution. 
%VAE has two part, the first one is encoder, which gets data as input and output latent vector, and the second one is decoder, which gets latent variable as input and output reconstructed data.
The objective function of VAE is represented as follows: 
\begin{equation}
    \log(p(x)) = D_{KL}(q_\phi(z|x)||p_\theta(z|x)) + \mathcal{L}(\theta, \phi, x)
\end{equation}
\begin{align} \label{eq:vae_2}
    \mathcal{L}(\theta, \phi, x) = - D_{KL}(q_\phi(z&|x)||p_\theta(z)) \notag\\ 
    &+ E_{z\sim q_\phi(z|x)}[\log(p_\theta(x|z))]
\end{align}
where $log(p(x))$ denotes the marginal log likelihood of the data, $\mathcal{L}(\theta, \phi, x)$ denotes the variational lower bound of marginal likelihood, $p_{\theta}(x|z)$ denotes the output distribution of the decoder, $q_\phi(z|x)$ denotes the output distribution of the encoder, and $p_{\theta}(z)$ denotes a normal distribution. By maximizing the lower bound, marginal likelihood of data is maximized. In VAE, latent space is assumed as a multivariate Gaussian distribution.

Doersch~\shortcite{doersch2016tutorial} indicated that conditional Variational Autoencoder (cVAE) is specifically used to learn a multimodal distribution via class information. The basic idea of cVAE is similar to that of VAE, which aims to learn the distribution of data. However, the encoder and the decoder of cVAE take a class label as an additional input. 

In this study, we select a cVAE structure to learn class-specific data distributions. We use the encoder as the mapping function of adversarial examples to the latent space of legitimate images and the decoder as the reconstructor of images from the latent spaces. We confirm that the forwarding process via the encoder-decoder model effectively purifies adversarial noise from data.

\subsection{Adversarial Examples} % 혜미
An adversarial example is a sample that is designed to be misclassified by the target classifier by using intended noise that is not perceivable by humans.
Prior to a study by Goodfellow~\shortcite{fgsm}, attackers exploited the non-linearity of neural networks. However, the authors claimed that the cause of vulnerability to adversarial examples is a linear characteristic of neural networks and proposed the Fast Gradient Sign Method (FGSM) that uses the gradient from the objective function of neural networks.
% \begin{equation}
%   \tilde{x} = x + \varepsilon \cdot sign(\nabla_{x}J(w, x, l))
%   \label{eq:FGSM}
% \end{equation}
% where $\tilde{x}$ is the adversarial example and $x$ is the untainted input and $l$ is the true label of $x$; $\varepsilon$ decides range of adversarial perturbation noise which is applied to the image, and $J$ is the cost function to train the network.

Although FGSM is a fast algorithm to make adversarial attacks, it is easy to defend the one step gradient based approach. In order to overcome the problem, the iterative Fast Graident Sign Method (iFGSM) was proposed by Kurakin~\shortcite{kurakin2016adversarial}. The method optimizes adversarial noise in several steps with a small perturbation of an image which allows a more accurate attack. 

RAND+FGSM \cite{Tramer2017ensemble} is a new attack method that adds random Gaussian noise to an image, and computes FGSM with the perturbed image. Against the FGSM-based methods, we used targeted adversarial attacks because we assume that it is more difficult to defend targeted attacks than untargeted attacks \cite{xu2018structured}. In our experiment, we randomly chose target labels among classes with the exception of the true class.

The Carlini and Wagner (CW) attack suggested by Carlini~\shortcite{Carlini2017CWattack} is the most powerful attack method among existing methods. By solving the optimization problem in a gradient descent manner, adversarial perturbation is derived as follows:
\begin{align}
    \mathrm{minimize} ~~ & ||\delta||_p + c * f(x + \delta) \\
    \mathrm{subject ~ to} ~~ & x + \delta \in [0, 1]^n
    \label{eq:CW-attack}
\end{align}
where $f(x)$ denotes an objective function of a classifier and $\delta$ denotes perturbation added to image $x$. In this study, we used $p=2$ and created CW attack with open source software CleverHans\footnote{\href{https://github.com/tensorflow/cleverhans}{https://github.com/tensorflow/cleverhans}} by Papernot~\shortcite{papernot2018cleverhans} to verify whether our method can defend the attack. 
%FGSM RANDFGSM, IFGSM, CW
% C & W 

\section{Proposed Method}
%% proposed method
% dataset 가정
In this paper, we propose a VAE-based defense method that is coined as PuVAE to purify adversarial noise from data. We consider a dataset $\mathcal{X}_{\mathrm{data}}$ that consists of data instances $\mathbf{x}_{\mathrm{data}} \in \mathbb{R}^d$ where $d$ denotes the dimension of the data space. Corresponding class labels (one-hot vectors) are denoted by $\mathbf{y}_{\mathrm{data}} \in \mathbb{R}^c$ in a set of classes $C$ where $c$ is the number of classes. 

% target, attack 모델 가정, adv_ex 가정 (target/attack 이용해서)
We then consider a target classifier $M_t$ that is the model an attacker wants to attack. 
%An attack classifier $C_a$ which is the model the attacker actually uses to attack. When the attacker uses the architecture and parameter of the target classifier, it is a white-box attack, otherwise it is a black-box attack. 
We also assume a set $\mathcal{X}_{\mathrm{adv}}$ that consists of adversarial examples $\mathbf{x}_{\mathrm{adv}} \in \mathbb{R}^d$ created from the target classifier. We define a set $\mathcal{X}$ which contains clean samples and adversarial examples. Instances $\mathbf{x}$ from the set $\mathcal{X}$ are used at inference time.
We explain the procedures of training and generating purified samples using PuVAE. The overview of the proposed method is described in Figure \ref{fig:model}.

%% PuVAE model
\subsection{Training process of PuVAE}
% conditional, dilated conv, epsilon sampling
%PuVAE model is comprised of an encoder and a decoder networks. The encoder는 conditional VAE와 같이 데이터 샘플과 label pair를 입력받아 입력된 label에 해당하는 latent space 상의 Gaussian distribution의 mean과 std를 출력한다. 우리는 dilated convolutional neural network을 encoder로 이용하였는데, conv에 비해 receptive field가 넓어지고 띄엄띄엄 보면서 local 한 정보를 잊어버리게 되기 때문에 layer를 거칠수록 error가 accumulate 되는 현상이 약해진다.
PuVAE is comprised of an encoder and a decoder network. The encoder receives a data-label pair and outputs the mean $\boldsymbol{\mu}$ and the standard deviation $\boldsymbol{\sigma}$ of the Gaussian distribution on the latent space corresponding to the input label:
\begin{align}
    \boldsymbol{\mu}, \boldsymbol{\sigma} &= \mathrm{Encoder}(\mathbf{x}_{\mathrm{data}},\mathbf{y}_{\mathrm{data}}) \label{eq:5}
\end{align}
Using $\mathbf{\mu}$ and $\mathbf{\sigma}$ obtained from the encoder, the latent vector $\mathbf{z}$ on the latent space is sampled:
\begin{align}
    \mathbf{z} &= \boldsymbol{\mu} + \boldsymbol{\epsilon} \cdot \boldsymbol{\sigma} \label{eq:6}\\
    \boldsymbol{\epsilon} &\sim N(\mathbf{0},\sigma_\epsilon \mathbf{I}) \label{eq:7}
\end{align}
where $\boldsymbol{\epsilon}$ denotes a random variable for the reparameterization trick, and $\sigma_\epsilon$ denotes a hyperparameter that is multiplied by the standard deviation to control the extent to which the latent vector is sampled. In the experiments, we used $\sigma_\epsilon=1$ in the training time to ensure that the posterior latent distribution follows the normal distribution.

In classification tasks, Convolutional Neural Networks (CNNs) using pooling and strides are used to select useful features and to widen receptive field. However, this selective nature of CNNs is a disadvantage on generative models, since the feature selection causes information loss. Therefore, we use a dilated convolutional neural network as the encoder to get the latent vector $\mathbf{z}$. Dilated convolution inserts zeros in the filter, so that the receptive field is enlarged and information loss is effectively reduced.

the sampled $\mathbf{z}$ enters the decoder with the label and produces an output instance $\hat{\mathbf{x}}$ with the same dimension $d$ as the input:
\begin{align}
    \hat{\mathbf{x}} = \mathrm{Decoder}(\mathbf{z}, \mathbf{y}_{\mathrm{data}})
\end{align}

At the training time, PuVAE is trained to maximize the variational lower bound in a manner similar to cVAE. Loss functions from the encoder and the decoder are:
\begin{align}
    \mathcal{L}_{\mathrm{RC}} = & ~ \mathbf{x}_{\mathrm{data}} \log \hat{\mathbf{x}} + (1-\mathbf{x}_{\mathrm{data}}) \log (1-\hat{\mathbf{x}})\\
    & \mathcal{L}_{\mathrm{KL}} = \boldsymbol{\mu}^2 + \boldsymbol{\sigma}^2 - \log (\boldsymbol{\sigma}^2-1)
\end{align}
where $\mathcal{L}_{\mathrm{RC}}$ denotes the reconstruction loss function to minimize the difference between the input instance and output instance, and $\mathcal{L}_{\mathrm{KL}}$ denotes Kullback-Leibler divergence between the output distribution of the encoder and the normal distribution. This process allows PuVAE to construct the mapping of legitimate data on the latent space
%With the process, we expect that the PuVAE constructs the mapping of legitimate data on the latent spaces. %Because PuVAE only learns the distribution of $\mathbf{z}$ from the training data, the input data is mapped to the learned latent spaces even if adversarial example comes in. Therefore, it can be assumed that adversarial noise will be removed in the projection to the latent variable.

% source model
Additionally, we use the cross-entropy calculated from a classifier as a loss function for PuVAE. The classifier, called \textit{source} classifier $M_s$, learns the decision boundaries on the data space. Since neural networks performing the same task learn similar functions \cite{fgsm}, we use a fixed architecture for $M_s$. Then, trained $M_s$ is used to ensure that the output instance reflects the characteristic of the classes in $C$. The cross-entropy loss from $M_s$ is as follows:
\begin{align}
    \mathbf{y}_s &= M_{s}(\hat{\mathbf{x}}) \\
    \mathcal{L}_{\mathrm{CE}} = \mathbf{y}_{\mathrm{data}} \log \mathbf{y}_s &+ (1-\mathbf{y}_{\mathrm{data}}) \log (1-\mathbf{y}_s)
\end{align} 
Finally, PuVAE is trained using the stochastic gradient descent (SGD):
\begin{align}\label{PuVAE_loss}
    \nabla ({\lambda}_{\mathrm{RC}}\mathcal{L}_{\mathrm{RC}} + {\lambda}_{\mathrm{KL}}\mathcal{L}_{\mathrm{KL}} + {\lambda}_{\mathrm{CE}}\mathcal{L}_{\mathrm{CE}})
\end{align}
where ${\lambda}_{\mathrm{KL}}$, ${\lambda}_{\mathrm{RC}}$, ${\lambda}_{\mathrm{CE}}$ are coefficients for each loss functions.

\begin{table}[t!]
\centering
\caption{Neural network architectures for PuVAE.}
\label{table:architecture}
{\resizebox{0.71\columnwidth}{!}
{\begin{tabular*}{0.87\columnwidth}{c|c}

\hline
	\toprule
   Encoder & Decoder \tabularnewline
   \midrule
    Dilated Conv(32, 7$\times$7, 2) & FC(512) \tabularnewline
    ReLU & ReLU \tabularnewline
    Dilated Conv(32, 7$\times$7, 2) & Deconv(32, 7$\times$7, 2) \tabularnewline
    ReLU & ReLU \tabularnewline
    Dilated Conv(32, 7$\times$7, 2) & Deconv(32, 7$\times$7, 2) \tabularnewline
    ReLU &  ReLU \tabularnewline
    FC(1024) & Deconv(32, 7$\times$7, 2) \tabularnewline
    ReLU &  Sigmoid \tabularnewline
    FC(1024)  &  \tabularnewline
    ReLU &  \tabularnewline
    FC(64) & \tabularnewline
    Softplus (for $\sigma$ only) & \tabularnewline
    &  \tabularnewline
    \bottomrule
\end{tabular*}}}
%}
\end{table}

%% PuVAE inference
% likelihood
\subsection{Generating Purified Samples}
% eps는 0.1을 썼다고만
At the inference time, PuVAE projects an input sample to the data manifolds of all classes in $C$ as follows:
\begin{equation} \label{PuVAE_1}
    \hat{\mathbf{x}}_{\mathbf{y}_i} = \mathrm{PuVAE}(\mathbf{x}, \mathbf{y}_i) 
\end{equation}
where $\mathbf{y}_i$ denotes the $i$-th class label in $C$ to guide the input to the corresponding latent space, and $\hat{\mathbf{x}}_{\mathbf{y}_i}$ denotes a candidate for the purified sample. The inference also follows the Equations (\ref{eq:5}), (\ref{eq:6}), and (\ref{eq:7}) as in training, where $\sigma_\epsilon$ is used to sample the latent vector $\mathbf{z}$. We performed a hyperparameter search on $\sigma_\epsilon$ among $\{0, 0.01, 0.1, 1, 10, 100\}$ for the inference. Thus, the optimal value of $\sigma_\epsilon$ is 0.1.

Because PuVAE only learns the distribution of $\mathbf{z}$ from the training data, the input data is mapped to the learned latent spaces even if adversarial example comes in. the adversarial noise is removed in the projection to the latent variable.

Then, the class label corresponding to the closest projection, $\mathbf{y}^*$, is selected as follows:
\begin{equation} \label{PuVAE_2}
    \mathbf{y}^* = \argmin_{\mathbf{y}_i \in C}~{D(\mathbf{x},\hat{\mathbf{x}}_{\mathbf{y}_i})}
\end{equation}
where $D$ denotes a distance measure to determine the closest projection. We use the root mean square error (RMSE) as the distance measure. Therefore, the candidate generated with label $\mathbf{y}^*$ is the purified sample which goes into $M_t$:
\begin{equation}
    \mathbf{x}_{\mathrm{purified}} = \hat{\mathbf{x}}_{\mathbf{y}^*}
\end{equation} 
Finally, the purified sample is fed into the target classifier $M_t$ as follows:
\begin{equation}
    \mathbf{y}_t = M_t(\mathbf{x}_{\mathrm{purified}})
\end{equation}
The complete process of generating the purified sample using PuVAE is illustrated in Figure~\ref{fig:PuVAE_model}.

\begin{figure}[t!]
\centering
\includegraphics[width=\columnwidth]{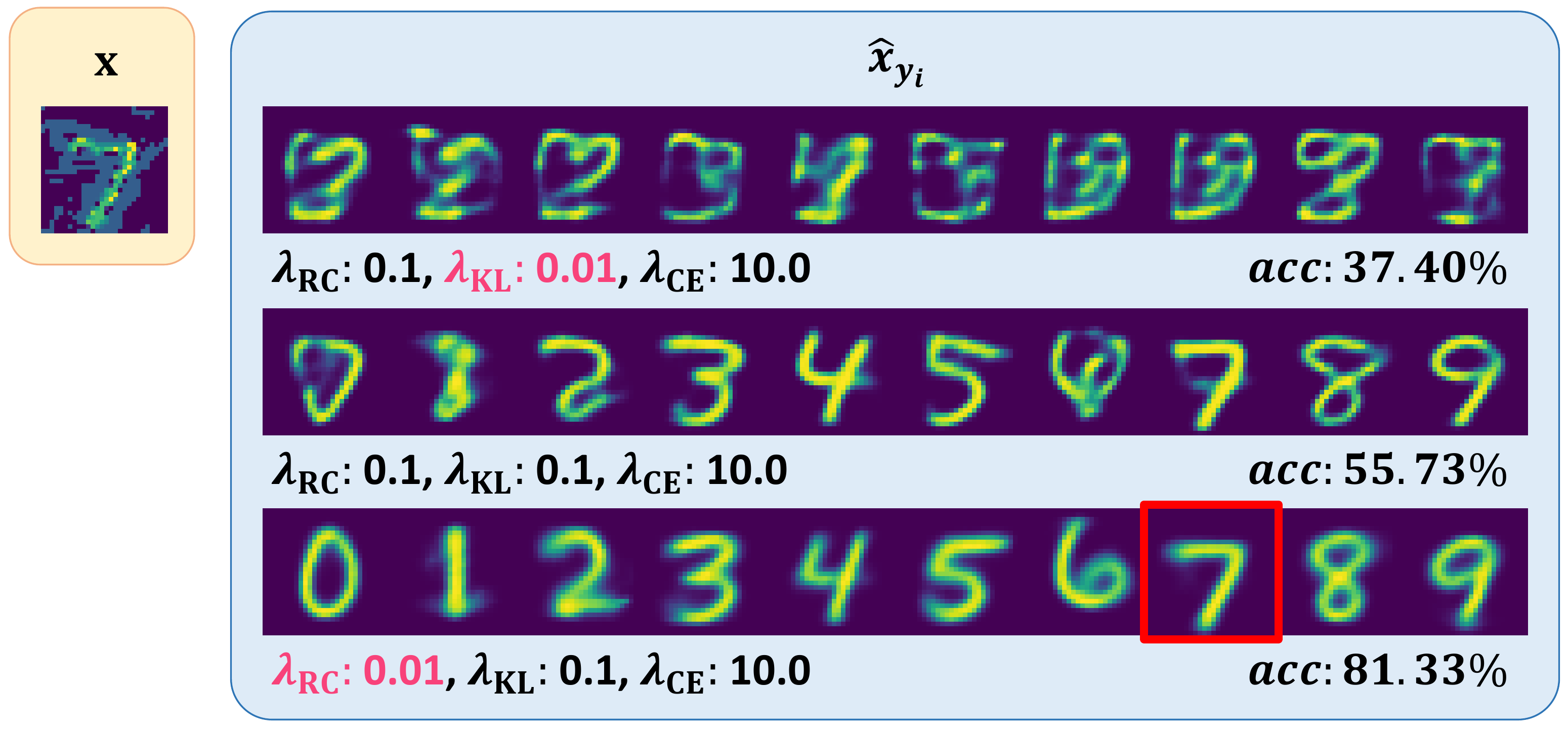}
\caption{Comparison of coefficients for training PuVAE; an adversarial example is highlighted with the yellow box
%We illustrate an adversarial example in the yellow box. 
The blue box represents the output samples when the input $\mathbf{x}$ enters. Each row shows output images from the model trained in different coefficient combinations. Each column shows images sequentially conditioned on class labels $0\sim9$. $acc$ denotes the defense performance from $M_t$ using the coefficient combination of each row.}
\label{fig:hyperparams}
\end{figure}

% coeff for hyperparams
% epsilon
% measure

\section{Experiments}
\newcolumntype{C}[1]{>{\centering}m{#1}}
\begin{table*}[t!]
\centering
\caption{Defense performance on the MNIST dataset (\%).}
\label{table:MNIST}
{\resizebox{0.92\textwidth}{!}
{\begin{tabular}{c|c|c|c|c|c|c|c}

\hline
	\toprule
	Classifier & Attacks & No Attack & No Defense & Adv. Tr. & MagNet & Defense-GAN & PuVAE \tabularnewline \midrule
	A & FGSM (0.3) & 99.51 & 12.46 & 78.57 & 26.90 & \textbf{85.29} & 81.33 \tabularnewline
	& iFGSM (0.3) & 99.51 & 0.00 & 91.72 & 72.48 & 87.40 & \textbf{92.33} \tabularnewline
	& RAND+FGSM (0.05, 0.3) & 99.51 & 10.64 &  84.21 & 32.34 & \textbf{89.02} & 82.70 \tabularnewline
	& CW (100) & 99.51 & 0.43 & 18.80 & 18.80 & 90.04 & \textbf{90.80} \tabularnewline \midrule
    B & FGSM (0.3) & 99.29 & 28.85 & \textbf{88.49} & 71.70 & 86.03 & 88.25 \tabularnewline
	& iFGSM (0.3) & 99.29 & 0.04 & \textbf{93.56} & 86.29 & 88.55 & 92.25 \tabularnewline
	& RAND+FGSM (0.05, 0.3) & 99.29 & 6.42 & 88.44 & 41.66 & \textbf{89.76} & 85.03 \tabularnewline
	& CW (100) & 99.29 & 0.59 & 19.20 & 19.10 & 90.76 & \textbf{92.92} \tabularnewline
	
    \bottomrule
\end{tabular}}
}
\end{table*}

\begin{table*}[t!]
\centering
\caption{Defense performance on the Fashion-MNIST dataset (\%).}
\label{table:F-MNIST}
{\resizebox{0.92\textwidth}{!}
{\begin{tabular}{c|c|c|c|c|c|c|c}

\hline
	\toprule
	Classifier & Attacks & No Attack & No Defense & Adv. Tr. & MagNet & Defense-GAN & PuVAE \tabularnewline \midrule
	A & FGSM (0.3) & 93.46 & 4.06 & 11.51 & 11.71 & 56.67 & \textbf{59.18} \tabularnewline
	& iFGSM (0.3) & 93.46 & 0.00 & 50.31 & 34.70 & 65.11 & \textbf{72.51} \tabularnewline
	& RAND+FGSM (0.05, 0.3) & 93.46 & 1.60 & 8.72 & 6.96 & \textbf{59.52} & 45.13 \tabularnewline
	& CW (100) & 93.46 & 5.01 & 18.00 & 15.50 & 68.60 & \textbf{80.59} \tabularnewline \midrule
    B & FGSM (0.3) & 93.54 & 2.75 & 8.00 & 15.30 & \textbf{58.20} & 52.46 \tabularnewline
	& iFGSM (0.3) & 93.54 & 0.00 & 46.97 & 41.84 & 66.06 & \textbf{71.04} \tabularnewline
	& RAND+FGSM (0.05, 0.3) & 93.54 & 1.34 & 9.93 & 7.44 & \textbf{60.98} & 52.22 \tabularnewline
	& CW (100) & 93.54 & 4.87 & 16.70 & 16.70 & 67.62 & \textbf{79.42} 
	\tabularnewline
    \bottomrule
\end{tabular}}
}
\end{table*}

\begin{table*}[t!]
\centering
\caption{Defense performance on the CIFAR-10 dataset (\%).}
\label{table:Cifar-10}
{\resizebox{0.92\textwidth}{!}
{\begin{tabular}{c|c|c|c|c|c|c|c}

\hline
	\toprule
	Classifier & Attacks & No Attack & No Defense & Adv. Tr. & MagNet & Defense-GAN & PuVAE \tabularnewline \midrule
	C &	FGSM (0.06)	& 82.13 &	3.82	&	19.07	&	18.58	&	29.74	&	\textbf{33.71} \tabularnewline
    	&	iFGSM (0.06)	& 82.13 &	0.32	&	24.13	&	27.00	&	33.69	&	\textbf{35.49} \tabularnewline
    	&	RAND+FGSM (0.005, 0.06)	& 82.13 &	4.84	&	21.27	&	19.99	&	\textbf{36.74}	&	33.76 \tabularnewline
    	&	CW (100)	& 82.13 &	9.88	&	\textbf{56.48}	&	40.13	&	38.13	&	36.36 \tabularnewline \midrule
    	D &	FGSM (0.06)	& 80.30 &	3.19	&	15.00	&	18.98	&	30.11	&	\textbf{33.20} \tabularnewline
    	&	iFGSM (0.06)	& 80.30 &	0.43	&	21.51	&	29.95	&	32.47	&	\textbf{34.48} \tabularnewline
    	&	RAND+FGSM (0.005, 0.06)	& 80.30 &	4.13	&	18.55	&	20.44	&	\textbf{35.76}	&	34.40 \tabularnewline
    	&	CW (100)	& 80.30 &	9.92	&	13.64	&	15.71	&	28.10	&	\textbf{31.70} \tabularnewline
    \bottomrule
\end{tabular}}
}
\end{table*}

%% dataset
% 이 섹션에서 우리는 optimal한 성능을 갖는 PuVAE와 Masked PuVAE를 찾고 이를 다른 방법들과 비교한다. 우리는 딥러닝에서 많이 이용되는 mnist dataset을 이용하여 실험을 수행하였다. mnist dataset은 hand written digit data로 50000개의 training set과 10000개의 test set으로 이루어져 있다. 우리는 각각의 mnist image를 0에서 1 사이의 값으로 scaling하여 학습과 inference에 이용하였다.
In this section, we determine optimal setting for PuVAE, and present the defense performance of PuVAE against adversarial attacks. We used Tensorflow (1.12.0) for the experiments. A GPU, an NVIDIA TITAN V (12 GB), and a CPU, an Intel Xeon E5-2690 v4 (2.6 GHz), were used. 
We used MNIST \cite{mnist} which is a hand-written digit dataset, Fashion-MNIST \cite{fmnist} which is a clothing object image dataset, and CIFAR-10 \cite{cifar10} which is a tiny natural image dataset. Each dataset consists of 50,000 training instances and 10,000 test instances. We normalized data between 0 and 1.

% attacks
We used FGSM, iFGSM, RAND+FGSM, and CW attacks for the experiments. FGSM, iFGSM, and RAND+FGSM were generated with an adversarial perturbation size of 0.3 for the MNIST and Fashion-MNIST datasets, and 0.06 for the CIFAR-10 dataset. We set the upper limit on the random noise of RAND+FGSM as 0.05 for the MNIST and the Fashion-MNIST datasets, and 0.005 for the CIFAR-10 dataset. We set the number of iterations of the CW attack as 100 on all datasets. The performance of defense mechanisms is measured by the accuracy of the target classifier. 

The architectures of the encoder and the decoder of PuVAE are presented in Table \ref{table:architecture}. Dilated Conv($n$, $k \times k$, $r$) denotes a dilated convolution layer with $n$ feature maps, filter size $k \times k$, and dilation rate $r$. Deconv($n$, $k \times k$, $s$) denotes a deconvolution layer with $n$ feature maps, filter size $k \times k$, and stride $s$. FC($m$) denotes a fully connected layer with $m$ units. ReLU denotes the rectified linear unit. We use the first half of the last layer of the encoder, 32 output units, as $\mu$ and the second half is passed to the softplus function to infer $\sigma$. We used the architecture of Defense-GAN and the reformer network of MagNet as suggested in \cite{defensegan} and \cite{magnet} respectively. The architectures of $M_t$ and $M_s$ are shown in the Supplementary Materials\footnote{\href{https://anonymous-puvae.github.io/}{https://anonymous-puvae.github.io/}}.

%% 준비과정 및 cVAE 학습
% 우리는 먼저 target classifier를 학습시킨 뒤 이로부터 adversarial examples와 saliency-map을 생성하였다. adversarial attack methods로는 FGSM과 iFGSM을 이용하였고, saliency-map은 주요 영역을 더 잘 캐치하도록 하기 위해 threshold 이상의 값은 1, 이하의 값은 0으로 만드는 binarize를 거쳤다.
% We first train the target classifier using training set. Then we create saliency-maps for each class label. Saliency-maps are binarized with a threshold of 0.25 to better capture key areas. We make adversarial examples from test set using FGSM and iFGSM as adversarial attack methods, and use test set as clean data.

% 우리는 conditional VAE를 training set으로 학습시켰다. PuVAE의 경우 input sample과 class label이 들어가서 비슷한 output sample이 생성되도록 reconstruction loss가 구성되고, Masked PuVAE의 경우 input sample이 masking되어 class label과 함께 들어가서 masking 이전의 input sample과 유사한 output sample이 생성되도록 reconstruction loss가 구성된다. 그리고 각 method의 성능은 target classfier의 classification accuracy로 측정하였다.
% Then we train conditional VAE with the training set. In the case of PuVAE, reconstruction loss is constructed so that an generated sample is similar to the input sample. In the case of Masked PuVAE, reconstruction loss is set to make generated sample be similar to the input sample before masking.

\subsection{Effect of coefficients on Training PuVAE}
%% 그림3
% 그림 3에서는, PuVAE의 특성을 알아보기 위해 학습시에 $w_recon$, $w_KL$, and $w_CE$ 세가지 loss에 대한 hyperparameter를 바꿔가며 generated samples의 특성을 알아보고, inference 성능을 측정하였다. 또 Masked PuVAE에 대해서도 노란 음영에 있는 an adversarial example과 초록 음영에 있는 0~9까지의 mask를 씌운 이미지, 그리고 파란 음영에 있는 generated samples를 그려보고 각각의 성능을 측정해 보았다.

% 우리는 학습시에 epsilon 1을 이용하였고 adversarial example이 입력으로 들어올 것에 대비하여 inference에 이용하기 위해 epsilon range (0, 0.01, 0.1, 1, 10, 100) 에 대하여 hyperparameter search를 진행하였다. 그 결과, 최적의 parameter가 epsilon 0.1을 이용하였다.
Figure \ref{fig:hyperparams} demonstrates the characteristics of generated samples based on the combinations of three coefficients $\lambda_{\mathrm{RC}}$, $\lambda_{\mathrm{KL}}$, and $\lambda_{\mathrm{CE}}$.
%We investigate the characteristics of generated samples in Figure \ref{fig:hyperparams} based on the combinations of three coefficients $\lambda_{\mathrm{RC}}$, $\lambda_{\mathrm{KL}}$, and $\lambda_{\mathrm{CE}}$. 
% RC
If $\lambda_{\mathrm{RC}}$ is larger than $\lambda_{\mathrm{KL}}$, the constraint of the posterior distribution of the encoder is relieved. 
Thus, the encoder easily maps the input samples to the low likelihood area of the latent space. The characteristic of this mapping causes the decoder to generate strange image as demonstrated in the first row of Figure \ref{fig:hyperparams}. 
%the input sample is mapped strictly to the normal distribution, whereas if $w_{\mathrm{RC}}$ is high, the influence of $\mathcal{L}_{\mathrm{KL}}$ becomes small, so  If a sample that has not been seen in the training time enters, it is mapped to $\mathbf{z}$ in the latent space out of the decoder's coverage, and a weird sample is generated by the decoder. When an adversarial example that has different distribution with the training data comes in, PuVAE generates an image of a strange shape as shown in the first row of the Figure \ref{fig:hyperparams}. 

% KL
Conversely, the typical form of each class is generated when $\lambda_{\mathrm{KL}}$ is large in model learning. Therefore, as shown in the last row of Figure \ref{fig:hyperparams}, it is possible to generate samples that exhibit distinctive characteristics of each class even if an input sample in other classes come in. The red box of Figure \ref{fig:hyperparams} illustrate that the purified sample $\mathbf{x}_{\mathrm{purified}}$ is analogous to the input sample $\mathbf{x}$, with the effect of adversarial noise removed. Additionally, the highest defense performance is acquired from the coefficient combination in the last row of Figure \ref{fig:hyperparams}. Therefore, we set $\lambda_{\mathrm{RC}}$, $\lambda_{\mathrm{KL}}$ and $\lambda_{\mathrm{CE}}$ to 0.01, 0.1, and 10, respectively, as coefficients in experiments.

\subsection{Defense Performance}
In this section, we compare the defense performance of PuVAE with maximum defense abilities of adversarial training, MagNet, and Defense-GAN. To obtain the best performance of Defense-GAN, we set the number of iterations as 200 and the number of candidates as 20. Tables \ref{table:MNIST}, \ref{table:F-MNIST}, and \ref{table:Cifar-10} show the performances of defense methods on the MNIST, the Fashion-MNIST, and the CIFAR-10 datasets respectively. 

As shown in Table \ref{table:MNIST}, the performance of PuVAE exceeds that of MagNet on all attacks and is comparable to that of Defense-GAN. Adversarial training is also comparable with our method in FGSM, iFGSM, and RAND+FGSM albeit a very low performance in the CW attack. Since we used gradients from $M_t$ for adversarial training, it is robust against gradient based attacks (FGSM, iFGSM, and RAND+FGSM), but weak against the other attack.

%As shown in Table \ref{table:F-MNIST}, Fashion-MNIST dataset에서는 두가지 architecture에서 모두 iFGSM과 CW attacks에 대해 PuVAE가 가장 좋은 성능을 보였다. 이는 iFGSM과 CW가 가장 defense하기 어려운 공격임에도 불구하고 the proposed model이 효과적으로 방어함을 demonstrate한다.
As shown in Table \ref{table:F-MNIST}, PuVAE shows the best performance against iFGSM and the CW attacks in both architectures. Even though iFGSM is the most difficult attack when there is no defense, the proposed model effectively defends the attack. The purified images on the MNIST and the Fasion-MNIST datsets are visualized in Supplementary Materials.

% In Table \ref{table:Cifar-10}, CIFAR-10 dataset에서는 전체적으로 낮은 정확도를 보였다. Adversarial training은 특정 setting에서 좋은 성능을 보였지만, model과 attack에 dependent한 결과를 보였다. 그러나 PuVAE는 다양한 attacks와 model architectures에서 가장 좋은 성능을 보인다. 또한 가장 좋은 성능을 보이지 못한 setting에서도 꽤 훌륭한 성능을 보인다. 이는 our method가 가장 general한 defense ability를 가졌음을 의미한다. 
As shown in Table \ref{table:Cifar-10}, the CIFAR-10 dataset shows overall low accuracy. Although adversarial training exhibits the best performance at a certain setting, it shows unstable results depending on models and attacks. However, PuVAE shows the best performance in various attacks and model architectures. Our method also exhibits a robust performance in settings where it does not take the first place. This indicates that the proposed method shows the general defense ability across various attacks.

\subsection{Performance in a Reasonable Time Constraint}
% 시간제한을 둔 후에 성능비교 그래프
%Defense-GAN, PuVAE와 같은 Purifying model의 경우 target classifier의 input을 preprocessing하는 형태를 띄고 있다. 최근들어 자율 주행에서 adversarial example이 security issue를 만들 수 있다는 주장이 펼쳐지고 있다. 이를 효과적으로 막을 수 있는 방법론이 필요한데 만약 purifying model을 자율 주행과 같이 real-time application에 사용할 경우 preprocessing 과정에 상당한 시간이 소요되는 것은 큰 문제가 된다. 따라서 우리는 real-time을 기준으로 성능을 측정하였고, 이는 그래프~ 에서 확인할 수 있다. PuVAE는 한 번의 inference로 정화된 샘플을 만들어내기 때문에 real-time에서도 좋은 성능을 보이는 반면, defense-gan은 최적화 과정을 거쳐 optimal한 hidden vector를 만들어야 하므로 시간제한 안에서 몇 \%의 낮은 성능을 보였다. PuVAE는 defenseGAN에 비해 정화샘플을 generate하는데 드는 시간을 효율적으로 줄인다. 

\begin{figure}[t!]
\centering
\includegraphics[width=1\columnwidth]{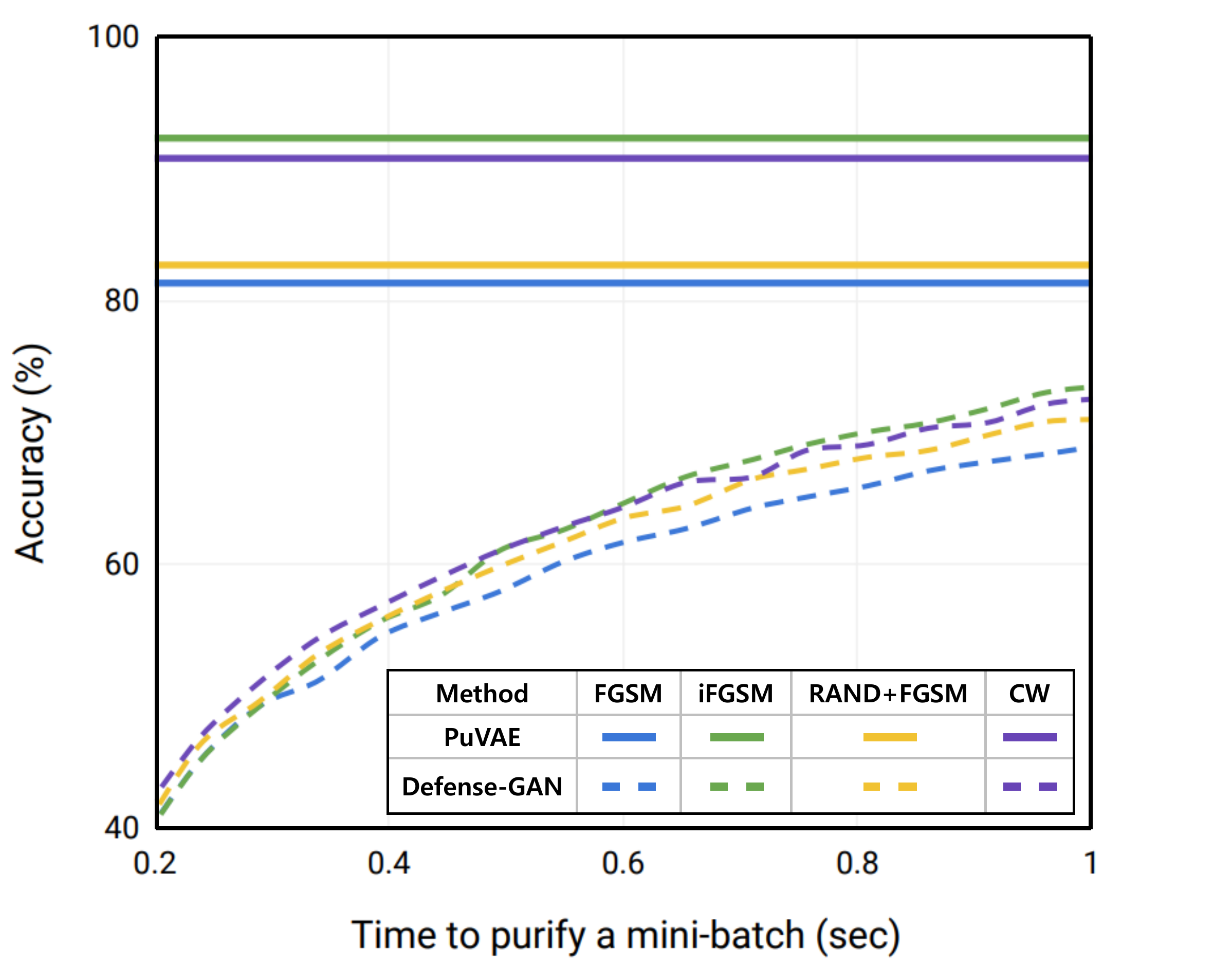}
\caption{Comparison with Defense-GAN in a reasonable time constraint.}
\label{fig:real-time}
\end{figure}

Defense mechanisms including Defense-GAN and PuVAE purify adversarial examples in a pre-processing manner. In contrast to PuVAE, Defense-GAN takes a significant amount of time to derive the maximum performance. While Defense-GAN takes approximately 14.8 seconds, PuVAE takes 0.114 seconds to purify a mini-batch with 128 MNIST images, allowing nearly 130 times faster inference as shown in Table \ref{table:inference_time}.

The security issue of adversarial attacks becomes particularly prominent in autonomous driving because it is not plausible to apply purifying-models that take a considerable amount of time to real-time applications. This setback of Defense-GAN accentuates the need for time-efficient defense methods.
Therefore, we measured the performance based on a reasonable time limit with the MNIST dataset. In the experiments, we set the time limit as one second.

In Figure \ref{fig:real-time}, the solid lines show the performances of PuVAE, and the dotted lines show the performances of Defense-GAN. Each color denotes a different attack method. PuVAE performs with one inference, and thus the performance of PuVAE is superior to that of Defense-GAN within the time limit. Since Defense-GAN creates a hidden vector iteratively by the gradient-based optimization process, the performance increases as time lapses. However, its performance does not reach the performance of PuVAE in the time limit. Therefore, PuVAE is more efficient than the state-of-the-art method for real-time applications.

%표 5에서, 우리는 Defense-GAN이 우세한 setting, where we used target classifier A and MNIST dataset, the reasonable time limit을 설정하였을 때 각 defense methods의 성능을 측정하였다. 그런데 DefenseGAN의 inference time이 PuVAE보다 훨씬 길기 때문에 불공정한 비교라고 생각했다. 그래서 우리는 1초의 time limit을 주고 defense methods를 공정하게 비교하였다. 그 결과 Defense-GAN의 성능이 maximum performance에 한참 미치지 못하는 것을 확인할 수 있었다. 따라서 reasonable한 time condition 내에서는 우리의 proposed method가 가장 높은 성능을 나타내므로 많은 real-world application에 활용될 수 있을 것으로 기대된다.
It is unfair to compare PuVAE and Defense-GAN without time constraint because the inference time of Defense-GAN significantly exceeds that of PuVAE. We compared the two defense methods after setting the time limit to one second. As shown in Table \ref{table:time_constraint}, it is observed that the performance of Defense-GAN is significantly lower than its maximum performance. Therefore, PuVAE is more practical in real-world scenarios because it exhibits the highest performance in a reasonable time condition.

\begin{table}[t!]
\centering
\caption{Inference time comparison of PuVAE and Defense-GAN.}
\label{table:inference_time}
{\resizebox{0.75\columnwidth}{!}
{\begin{tabular}{c|c|c}

\hline
	\toprule
	Method & PuVAE & Defense-GAN
	\tabularnewline \midrule
	Time (seconds) & \textbf{0.11} & 14.80
	\tabularnewline
	Ratio & 1 & 134.55
	\tabularnewline
    \bottomrule
\end{tabular}}
}
\end{table}

\begin{table}[t!]
\centering
\caption{Accuracy of defense methods within one second (\%).}
\label{table:time_constraint}
{\resizebox{0.98\columnwidth}{!}
{\begin{tabular}{c|c|c|c|c}

\hline
	\toprule
	Method & FGSM & iFGSM & RAND+FGSM & CW \tabularnewline \midrule
	Adv. Tr. & 78.57 & 91.72 & 84.21 & 18.80 
	\tabularnewline
	MagNet & 26.90 & 72.48 & 32.34 & 18.80
	\tabularnewline
	Defense-GAN & 69.25 & 73.85 & 71.50 & 72.79
	\tabularnewline
	PuVAE & \textbf{81.33} & \textbf{92.33} & \textbf{82.70} & \textbf{90.80}
	\tabularnewline
    \bottomrule
\end{tabular}}
}
\end{table}

% \subsection{Distance Measure}
% %To measure the distance between input and output of cVAE, we select various types of distance measure such as principal component vector cosine similarity, Struce similarity, feature map distance between convolution layer and RMSE.
% To verify that our distance measure works properly, we visualized and analyzed a case that is erroneously purified. As shown in the Figure \ref{fig:distance_comparison}, we can see that the right sample is similar to the middle sample with the naked eyes. However, the our distance measure, RMSE, gives a closer distance between the left sample and the intermediate sample than distance between the right sample and the middle sample. Therefore, we expect the performance of PuVAE could be improved by looking for a robust distance measure for parallel movement and distortion in the future.
% dataset, 실험방법

\section{Conclusion}

%본 논문에서는 PuVAE를 이용함으로써 adversarial example의 영향력을 purify 할 수 있다는 것을 확인하였다. 또한, 다양한 공격에 robust하지 않은 adversarial training의 단점과 purified sample을 generate하는데 오랜시간이 걸린다는 defense-GAN의 단점을 해결하였다. 하지만 PuVAE의 성능이 defense-GAN의 성능을 넘어서지 못하였고 classifer 성능의 상한과 같은 역할을 하였다. 성능은 distance measure, mask,  generating ability of cVAE에 dependent하다. 그 중에서도 distance measure에 의해서 큰 성능 차를 보였다. 본 논문에서는 RMSE, SSIM 등 여러가지 measure에 대해서 실험하였지만 defense-GAN의 성능을 넘어서는 measure를 발견하지 못하였다. 향후 평행이동이나 이미지 특이점에 robust한 distance measure를 찾음으로써 PuVAE의 성능을 높일 수 있을 것이라 예상한다. 

In this paper, we propose PuVAE, a novel VAE-based defense method that effectively purifies adversarial attacks. PuVAE is robust against various attacks and overcomes the disadvantages of adversarial training. The performance of PuVAE is also comparable to the best performance of Defense-GAN. In addition, PuVAE significantly ourperforms Defense-GAN given a reasonable time limit. We demonstrate the advantages of the proposed method on various datasets and adversarial attacks. For future work, we plan to apply our method to real-time applications such as autonomous-driving, face identification, and surveillance systems.

%we expect the performance of PuVAE could be improved by looking for a distance measure which is robust to adversarial examples hard to being purified.

% \begin{figure}[t!]
% \centering
% \includegraphics[width=0.7\columnwidth]{fig/distance_comparison.pdf} 
% \caption{Comparison of distance between samples; The left and right images are samples generated from PuVAE. The left image is the closest to the intermediate clean sample and the right image is the sample generated using the same label as the intermediate image.}
% \label{fig:distance_comparison}
% \end{figure}

%\section*{Acknowledgments}

%\appendix

\vfill

\pagebreak
\balance
\bibliographystyle{named}
\bibliography{ijcai19}

\end{document}